\documentclass[conference]{IEEEtran}
\IEEEoverridecommandlockouts

\usepackage{cite}
\usepackage{amsmath,amssymb,amsfonts}
\usepackage{algorithm}
\usepackage{algorithmic}
\usepackage{graphicx}
\usepackage{textcomp}
\usepackage{xcolor}
\usepackage{booktabs}
\usepackage{multirow}
\usepackage{colortbl}
\usepackage{pifont}
\usepackage{url}
\usepackage{amsthm}
\usepackage{tikz}
\usetikzlibrary{arrows.meta,positioning,fit,backgrounds}

\definecolor{rowgray}{RGB}{245,245,245}
\definecolor{hilite}{RGB}{232,242,254}
\newcommand{\cmark}{\textcolor{black!75}{\ding{51}}}

\newcommand{\NAx}{\textcolor{black!45}{\textsc{n/a}}}
\newcommand{\fidp}{\ensuremath{\mathrm{Fid}^{+}}}
\newcommand{\fidm}{\ensuremath{\mathrm{Fid}^{-}}}
\newcommand{\charact}{\ensuremath{\mathrm{Char}}}
\newtheorem{definition}{Definition}   
\newtheorem{proposition}{Proposition}

\begin{document}

\title{Amortised Post-Hoc Explanation with Exact Preservation for Dynamic Graph Anomaly Detectors}

\author{\IEEEauthorblockN{Iyad Assaad Nekka,
 Hamida Seba,
 Khaled-Walid Hidouci,
 Karima Amrouche} \\
\IEEEauthorblockA{LCSI Laboratory, National higher School of
Computer Science (ESI), Algiers, Algeria}
\IEEEauthorblockA{Universit\'e Claude Bernard Lyon, Lyon, France} \\
\IEEEauthorblockA{\{i\_nekka, w\_hidouci, k\_amrouche\}@esi.dz,\quad
hamida.seba@univ-lyon1.fr}}

\maketitle

\begin{abstract}
Anomaly detection in dynamic graphs underpins high-stakes applications in financial
fraud analysis, intrusion detection, and online platform integrity, where
accountability frameworks increasingly require that automated decisions be
accompanied by human-interpretable justifications. StrGNN, the strongest performer
in recent unified benchmarks of the field, delivers this detection quality while
producing no explanation whatsoever: when an edge is flagged, the analyst receives a
score and nothing more. Explanation quality metrics are not low for StrGNN --- they
are \emph{undefined}, because no attribution vector exists to evaluate. This paper
closes that gap. We present X-StrGNN, a post-hoc explanation layer that wraps a
trained, frozen StrGNN and emits, for every flagged edge, two complementary
attributions: a \emph{structural} attribution over the enclosing-subgraph evidence
identifying which contextual interactions drove the decision, and a \emph{temporal}
attribution over the observation window identifying which historical snapshot
carried the signal. Because both attributions are realised as multiplicative masks
that are identically one in the unexplained pass, the layer is an exact
pass-through: detection is preserved to machine precision, which we verify rather
than assert ($\Delta$AUC $=0.0000$, $\Delta$AP $=0.0000$, $\Delta$P@100 $=0.0000$).
Attribution costs $0.66$\,ms per flagged edge, making explanation of an entire alarm
list feasible rather than a hand-selected handful. We further conduct the first
controlled design study of attribution strategies for this architecture, comparing
gradient attribution, per-instance mask optimisation, and amortised
parameterisation under one protocol, one budget, and three seeds. X-StrGNN attains
the highest attribution stability in the study ($0.913$) at $268\times$ lower cost
than per-instance optimisation, and its temporal attribution ($1.601$ against a
measured random floor of $0.973$) is separably better than its own ablated control,
while per-instance optimisation --- the most expensive strategy evaluated --- falls
\emph{below} that floor. Code, protocol, and per-seed measurements are released.
\end{abstract}


%=============================================================================
\section{Introduction}
%=============================================================================

Real-world networks evolve. Financial transaction graphs, enterprise network
traffic, online trust communities, and social platforms change continuously as
interactions appear and disappear. Detecting anomalous behaviour in these evolving
structures --- fraudulent transfers, lateral movement in a compromised network,
coordinated inauthentic activity --- is both practically urgent and technically
demanding~\cite{ekle2024survey}.

Deep learning has transformed detection quality. NetWalk~\cite{yu2018netwalk},
AddGraph~\cite{zheng2019addgraph}, StrGNN~\cite{cai2021strgnn} and
TADDY~\cite{liu2021taddy} exploit graph neural networks and attention to capture
structural and temporal dependencies jointly. Among these, StrGNN occupies a
distinctive position: it scores a candidate edge by extracting an $h$-hop enclosing
subgraph in each of $w$ consecutive snapshots, encoding each with a graph
convolutional stack and SortPooling~\cite{zhang2018dgcnn}, and passing the resulting
sequence through a gated recurrent unit. In recent unified benchmarking of the
subfield it leads the canonical injected benchmark.

There is, however, something StrGNN cannot do: explain itself. When edge $(u,v)$ is
flagged at snapshot $t$, the detector emits a scalar. It does not indicate which
contextual interactions in the enclosing subgraph constituted the evidence, nor
whether the anomaly crystallised at the current snapshot or had been accumulating
across the observation window. In fraud compliance, an analyst cannot act on an
alert they cannot interrogate. In regulated environments, a score without a
justification is operationally insufficient. This is not a shortcoming peculiar to
StrGNN; it is the default condition of essentially every deep dynamic-graph anomaly
detector catalogued in the literature~\cite{ekle2024survey}, and post-hoc
explainability for this setting has only recently begun to
emerge~\cite{nekka2026gram}.

This paper closes the gap for StrGNN. The contribution is not an incremental
fidelity improvement over an existing explainer for this architecture, because no
such explainer exists. Explanation quality metrics are not \emph{low} for StrGNN;
they are \emph{undefined}, since there is no attribution vector to evaluate. Our
framework supplies two, and does so at exactly zero cost to detection.

\noindent\textbf{Contributions.}
\begin{enumerate}
\item \textbf{The first post-hoc explanation layer for StrGNN}, producing
\emph{dual structural and temporal attribution} for every flagged edge over a frozen
detector, with detection preserved to machine precision ($\Delta$AUC $=0.0000$).
\item \textbf{An amortised spatio-temporal parameterisation} in which one shared
network emits both a structural mask over enclosing-subgraph messages and a gate
over the observation window, trained under separate counterfactual objectives, and
delivering attribution in $0.66$\,ms per edge --- $268\times$ faster than
per-instance optimisation.
\item \textbf{The first controlled design study of attribution strategies for this
architecture}, comparing four strategies against a measured random floor under one
protocol, one budget, and three seeds, yielding deployment-driven guidance.
\item \textbf{A documented audit of the released StrGNN implementation}, reporting
five defects that change published numbers, including a tensor-axis error under
which recurrent step $t$ does not correspond to snapshot $t$ --- which invalidates
any temporal attribution computed without the correction.
\end{enumerate}

%=============================================================================
\section{Related Work}
%=============================================================================

\subsection{Anomaly Detection in Dynamic Graphs}

Early approaches relied on sketches and handcrafted structural signatures, flagging
statistical outliers in edge streams at high throughput. Deep methods raised
detection quality substantially: NetWalk~\cite{yu2018netwalk} combined random-walk
embeddings with a reservoir; AddGraph~\cite{zheng2019addgraph} paired a GCN with an
attention-augmented GRU; StrGNN~\cite{cai2021strgnn} introduced enclosing-subgraph
extraction with double-radius node labelling~\cite{zhang2018seal};
TADDY~\cite{liu2021taddy} unified structural and temporal signal in a single
Transformer encoder; SLADE~\cite{lee2024slade} achieved constant per-edge cost in
the streaming setting. None addresses the interpretation of its own predictions.

\subsection{Post-Hoc Explanation for Graph Neural Networks}

GNNExplainer~\cite{ying2019gnnexplainer} optimises a soft edge mask per instance so
that the retained subgraph reproduces the prediction.
PGExplainer~\cite{luo2020pgexplainer} amortises that optimisation into a network
shared across instances, replacing per-instance search with a single forward pass
and rendering explanation inductive. Gradient attribution requires no training and
remains a standard reference point~\cite{yuan2022taxonomic}. All were designed for
static graphs, and none possesses an axis on which to express \emph{when} a decision
was made --- precisely the degree of freedom a windowed detector introduces.

\subsection{Explanation for Temporal Models}

TGNNExplainer~\cite{xia2023tgnnexplainer} searches over explanatory events for
continuous-time temporal GNNs, and TempME~\cite{chen2023tempme} identifies temporal
motifs; both target continuous-time event streams rather than the discrete-snapshot
window StrGNN consumes. Han et al.~\cite{han2026explainable} address dynamic
\emph{heterogeneous} graphs through relation-evolution patterns, a complementary
setting with a different input model. To our knowledge no prior work supplies
attribution for the enclosing-subgraph-plus-recurrence architecture, which is the
gap this paper fills.

%=============================================================================
\section{Preliminaries}
%=============================================================================

\begin{definition}[Dynamic graph]
A dynamic graph is a sequence $\mathcal{G}=\{G_1,\dots,G_T\}$ where each
$G_t=(V,E_t)$ shares a node set $V$ with a time-varying edge set $E_t$.
\end{definition}

\begin{definition}[Edge-level anomaly detection]
A scorer $f:(e,t)\mapsto[0,1]$ assigns each candidate edge $e=(u,v)$ at snapshot
$t$ a score; edges exceeding a threshold are flagged.
\end{definition}

\subsection{StrGNN}

StrGNN~\cite{cai2021strgnn} decomposes into three stages. \emph{Enclosing subgraph
generation} extracts, for target edge $(u,v)$ at snapshot $t$, the $h$-hop enclosing
subgraph in each of the snapshots $t-w+1,\dots,t$, and assigns double-radius node
labels~\cite{zhang2018seal} encoding each node's role relative to $u$ and $v$. The
target link is removed from its own subgraph, so the evidence is strictly
contextual. \emph{Structural feature extraction} applies a graph convolutional
stack~\cite{kipf2017gcn} and SortPooling to yield a fixed-size snapshot
representation $s_t$. \emph{Temporal detection} passes the length-$w$ sequence
$(s_{t-w+1},\dots,s_t)$ through a GRU and a classifier.

Two properties govern what an explanation of StrGNN must be. The evidence is a set
of contextual interactions rather than the target edge itself; and the decision is a
function of a \emph{sequence}, so any faithful explanation carries a temporal degree
of freedom that static attribution cannot express.

%=============================================================================
\section{The X-StrGNN Framework}
\label{sec:framework}
%=============================================================================

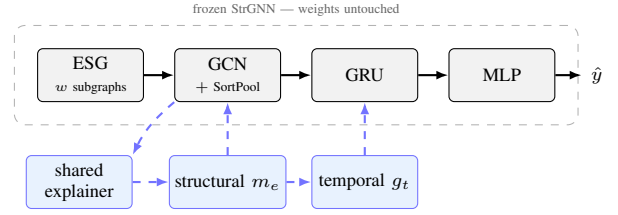
\begin{figure}[t]
\centering
\begin{tikzpicture}[
  font=\scriptsize,
  box/.style={draw, rounded corners=2pt, minimum height=7mm, minimum width=14mm,
              align=center, inner sep=2pt},
  frozen/.style={box, fill=black!5},
  learn/.style={box, fill=hilite, draw=blue!55},
  ar/.style={-{Latex[length=1.6mm]}, thick},
  dar/.style={-{Latex[length=1.6mm]}, thick, blue!55, dashed}]
\node[frozen] (esg) {ESG\\{\tiny $w$ subgraphs}};
\node[frozen, right=4mm of esg] (gcn) {GCN\\{\tiny $+$ SortPool}};
\node[frozen, right=4mm of gcn] (gru) {GRU};
\node[frozen, right=4mm of gru] (mlp) {MLP};
\node[right=3.5mm of mlp] (out) {$\hat{y}$};
\draw[ar] (esg)--(gcn); \draw[ar] (gcn)--(gru);
\draw[ar] (gru)--(mlp); \draw[ar] (mlp)--(out);
\node[learn, below=7mm of gcn] (me) {structural $m_e$};
\node[learn, below=7mm of gru] (gt) {temporal $g_t$};
\node[learn, below=7mm of esg, xshift=-1.5mm] (net) {shared\\explainer};
\draw[dar] (net)--(me); \draw[dar] (me)--(gt);
\draw[dar] (me)--(gcn); \draw[dar] (gt)--(gru);
\draw[dar] (gcn.south west) to[bend right=12] (net.north east);
\begin{scope}[on background layer]
\node[fit=(esg)(mlp), draw=black!30, dashed, rounded corners, inner sep=3mm,
      label={[font=\tiny,black!60]above:frozen StrGNN --- weights untouched}] {};
\end{scope}
\end{tikzpicture}
\caption{X-StrGNN. The detector (grey) is frozen; a shared explainer network (blue)
emits a structural mask over enclosing-subgraph messages and a gate over the $w$
recurrent inputs, conditioned on the detector's own internal representations. Both
masks are multiplicative and identically one in the unexplained pass, so the
composition reduces \emph{exactly} to StrGNN.}
\label{fig:arch}
\end{figure}

\subsection{Design Principles}

\noindent\textbf{P1: Post-hoc, zero modification.} The framework wraps a trained,
frozen detector. Weights, training procedure, and inference path are untouched, so
detection quality is preserved by construction rather than by empirical
verification --- though we verify it regardless (Table~\ref{tab:parity}).

\noindent\textbf{P2: Dual decomposition.} A single attribution vector over edges is
insufficient for a windowed detector. X-StrGNN (Fig.~\ref{fig:arch}) decomposes along two orthogonal axes:
\emph{structural} (which contextual interactions?) and \emph{temporal} (which
snapshot?). The two questions are distinct, and an analyst needs both.

\noindent\textbf{P3: Amortisation.} Explanation must be cheap enough to apply to an
entire alarm list. A shared network delivers attribution in one forward pass
(Algorithm~\ref{alg:xstrgnn}) and is inductive: edges unseen at explainer-training time are explained without
re-optimisation.

\subsection{Masked Forward Pass}

Let $f_\phi$ be the frozen detector and $\mathcal{G}$ the window of enclosing
subgraphs. We introduce $m_e\in[0,1]$ on the message carried by each undirected
enclosing-subgraph edge and $g_t\in[0,1]$ on each of the $w$ recurrent inputs. The
masked message aggregation at layer $\ell$ is
\begin{equation}
h^{(\ell)}_i = \tanh\!\Big( \mathbf{D}^{-1} W^{(\ell)}
\big( h^{(\ell-1)}_i + \textstyle\sum_{j\in\mathcal{N}(i)}
m_{ij}\, h^{(\ell-1)}_j \big) \Big),
\label{eq:masked}
\end{equation}
and the recurrent input at step $t$ becomes $g_t\cdot s_t$. Degree normalisation
$\mathbf{D}$ is computed on the \emph{unmasked} graph, which is what makes the
following identity exact.

\begin{proposition}[Exact pass-through]
$f_\phi(\mathcal{G}\mid m\!\equiv\!1,\,g\!\equiv\!1)=f_\phi(\mathcal{G})$
identically.
\end{proposition}

\noindent This is a structural guarantee rather than an empirical near-miss: both
masks enter multiplicatively and normalisation is mask-independent, so the
unexplained pass reduces term-by-term to the original computation.

\subsection{Amortised Parameterisation}

Following the parameterisation principle of PGExplainer~\cite{luo2020pgexplainer}, a
single network $\psi$ shared across all instances predicts mask logits from the
detector's own representations. For an edge $(i,j)$ in the subgraph of snapshot $t$
with target endpoints $(u,v)$,
\begin{equation}
\omega_{ij}=B\tanh\!\big(\mathrm{MLP}^{e}_\psi
([\,z_i\|z_j\|z_u\|z_v\|\rho_e(t)\,])/B\big),
\end{equation}
where $z$ are the concatenated graph-convolutional representations, $\rho_e$ a
learned positional embedding, and $B$ a bound preventing the relaxation from
saturating to a constant. Masks are sampled through the binary concrete
relaxation~\cite{maddison2017concrete,jang2017gumbel},
$m=\sigma((\omega+\log u-\log(1-u))/\tau)$ with $u\sim\mathcal{U}(0,1)$ and $\tau$
annealed from $5.0$ to $0.5$. A second head produces temporal logits from the
snapshot representations and their window context, using a \emph{separate}
positional embedding $\rho_\tau$; the two heads share no parameters, a choice
examined in Section~\ref{sec:designnotes}.

\subsection{Objective}

Writing $\hat{y}=\arg\max f_\phi(\mathcal{G})$ for the detector's own prediction,
\begin{align}
\mathcal{L}=\;& \mathrm{NLL}\big(f_\phi(\mathcal{G}\mid m),\hat{y}\big)
\;+\;\lambda_s\big[\kappa-\mathrm{NLL}\big(f_\phi(\mathcal{G}\mid 1-m),
\hat{y}\big)\big]_{+} \nonumber\\
&+\;\lambda_\tau\big[\kappa-\mathrm{NLL}\big(f_\phi(\mathcal{G}\mid
g^{\mathrm{abl}}),\hat{y}\big)\big]_{+}\;+\;\mathcal{R}.
\label{eq:obj}
\end{align}
The first term enforces \emph{sufficiency}: the retained evidence alone reproduces
the decision. The second and third enforce \emph{necessity} on each axis separately:
discarding the structural explanation, or ablating the nominated snapshot, must
destroy the decision. $\mathcal{R}$ collects a density term, a mask entropy term,
and a temporal smoothness term so the gate reads as an evolution rather than
isolated spikes.

The temporal ablation is constructed to induce competition among snapshots:
\begin{equation}
a=\mathrm{softmax}(\omega^{\tau}/\tau),\qquad
g^{\mathrm{abl}}_t = 1-a_t/\max_{t'}a_{t'} ,
\label{eq:tabl}
\end{equation}
whose weights sum to one, so raising one snapshot necessarily lowers the others.
Normalising by the maximum reproduces the evaluation probe exactly, which zeroes the
nominated snapshot and leaves the remainder at unity.

\begin{algorithm}[t]
\caption{X-StrGNN attribution for a flagged edge}
\label{alg:xstrgnn}
\begin{algorithmic}[1]
\REQUIRE Frozen StrGNN $f_\phi$; trained explainer $\psi$; edge $e^*$ at $t^*$;
budget $p$
\ENSURE $\boldsymbol{\phi}_{\mathrm{struct}}$,
$\boldsymbol{\phi}_{\mathrm{temp}}\in\mathbb{R}^{w}$
\STATE $\mathcal{G}\leftarrow$ enclosing subgraphs of $e^*$ over
$t^*\!-\!w\!+\!1,\dots,t^*$
\STATE $(Z,s_{1:w})\leftarrow f_\phi(\mathcal{G})$
       \COMMENT{internal representations, one forward pass}
\STATE $\boldsymbol{\phi}_{\mathrm{struct}}\leftarrow
       \sigma(\mathrm{MLP}^{e}_\psi(Z))$ \COMMENT{per-edge attribution}
\STATE $\boldsymbol{\phi}_{\mathrm{temp}}\leftarrow
       \sigma(\mathrm{MLP}^{\tau}_\psi(s_{1:w}))$
       \COMMENT{per-snapshot attribution}
\RETURN top-$p$ structural evidence and
        $\arg\max_t \boldsymbol{\phi}_{\mathrm{temp}}$
\end{algorithmic}
\end{algorithm}

\subsection{Explanation Quality Metrics}

Every metric below presupposes an attribution vector. StrGNN without an explainer
produces none, so all are \emph{undefined} for it --- not zero, not low, but
undefined.

Raw fidelity is a probability difference and therefore scales with how confident a
given detector happens to be, which makes it incomparable across detectors and even
across checkpoints. We normalise by the \emph{ablation range} $\Delta_{\max}$, the
largest drop the detector can express, measured by masking all evidence:
\begin{equation}
\fidp=\frac{\mathbb{E}[p_0-p_{\setminus S}]}{\Delta_{\max}},\qquad
\fidm=\frac{\mathbb{E}[p_0-p_{S}]}{\Delta_{\max}},
\end{equation}
where $p_0$ is the probability of the detector's own prediction, $p_S$ that obtained
retaining only the explanation, and $p_{\setminus S}$ that obtained removing it.
Characterisation is the harmonic mean
$\charact=2\fidp(1-\fidm)/(\fidp+1-\fidm)$. \emph{Temporal fidelity} is the drop
induced by ablating the nominated snapshot divided by that induced by ablating a
uniformly chosen one, so that $1$ is the random floor --- and we \emph{measure} that
floor rather than assuming it. \emph{Stability} is the mean Spearman correlation of
per-subgraph attribution rankings under a small input perturbation. \emph{Sparsity}
is $1-p$.

%=============================================================================
\section{Experiments}
%=============================================================================

\subsection{Setup}

We evaluate on UCI-Messages~\cite{opsahl2009clustering}, the canonical DGAD
benchmark: $1{,}899$ nodes and $59{,}835$ timestamped interactions over $59$
snapshots of $1{,}000$ interactions. Training uses the first half of the stream with
context-dependent negative sampling as specified by StrGNN; the test partition
receives $10\%$ uniformly injected anomalies following the protocol established by
NetWalk~\cite{yu2018netwalk}, giving $6{,}240$ training and $1{,}653$ test targets
at a test anomaly rate of $0.105$. We set $w=5$, $h=1$, SortPooling $k=22$, and cap
enclosing subgraphs at $20$ nodes per hop. The explainer uses budget $p=20\%$,
$\lambda_s=1$, $\lambda_\tau=3$, $\kappa=2$, and trains for $30$ epochs with Adam at
$3\times10^{-3}$, the detector frozen throughout.

Executing the released StrGNN artefact under a single protocol surfaced five defects
that change reported numbers (Table~\ref{tab:patches}); all are patched before any
measurement here. P4 deserves emphasis: the tensor entering the recurrent module is
channel-major, and the released code applies a reshape rather than a transpose,
interleaving channels into the time axis. Any temporal attribution computed without
this correction is measured against a scrambled axis.

\begin{table}[t]
\caption{Defects found in the released StrGNN implementation, all patched before
measurement. P4 is load-bearing for any temporal claim.}
\label{tab:patches}
\centering\footnotesize
\begin{tabular}{@{}cl@{}}
\toprule
\textbf{ID} & \textbf{Defect} \\
\midrule
\rowcolor{rowgray}
P1 & Evaluation reports AP and F1 on the \emph{normal} class \\
P2 & Target link retained in its own enclosing subgraph (leakage) \\
\rowcolor{rowgray}
P3 & Training and test use different temporal windows \\
P4 & GRU input reshaped, not transposed: step $t\neq$ snapshot $t$ \\
\rowcolor{rowgray}
P5 & Device handling hardcoded to CUDA \\
\bottomrule
\end{tabular}
\end{table}

\subsection{Detection Is Preserved Exactly}

\begin{table}[t]
\caption{Detection parity. The explanation layer is inactive at $m\!=\!g\!=\!1$, so
detection is unchanged; the residual on raw scores is float32 epsilon.}
\label{tab:parity}
\centering\footnotesize
\begin{tabular}{@{}lccc@{}}
\toprule
\textbf{Metric} & \textbf{StrGNN} & \textbf{X-StrGNN} & \textbf{$\Delta$} \\
\midrule
\rowcolor{rowgray}
AUC-ROC & 0.8749834853 & 0.8749834853 & \textbf{0.0000} \\
AP      & 0.5163393989 & 0.5163393989 & \textbf{0.0000} \\
\rowcolor{rowgray}
P@100   & 0.6600000000 & 0.6600000000 & \textbf{0.0000} \\
\midrule
\multicolumn{3}{@{}l}{max $|\Delta s|$ over all test edges}
& $1.19\mathrm{e}{-}07$ \\
\bottomrule
\end{tabular}
\end{table}

Table~\ref{tab:parity} confirms Proposition~1 numerically: every detection metric is
bit-identical and the largest per-edge deviation is float32 epsilon arising in the
exponential. Explainability is obtained at exactly zero detection cost, and every
subsequent measurement explains the same model the detection column describes.

\subsection{Explainability Introduced}

Table~\ref{tab:gap} is the central result. Before this work no attribution exists
for StrGNN and every explanation metric is undefined.

\begin{table}[t]
\caption{Capabilities introduced. \NAx{} denotes \emph{undefined}: no attribution
vector exists for StrGNN, so evaluation is not possible. Mean over three seeds;
$\fidp$/$\fidm$ normalised by the detector ablation range ($0.326$).}
\label{tab:gap}
\centering\footnotesize
\begin{tabular}{@{}lccl@{}}
\toprule
\textbf{Capability / Metric} & \textbf{StrGNN} & \textbf{X-StrGNN} &
\textbf{Status} \\
\midrule
\rowcolor{rowgray}
Structural attribution & \NAx & \cmark & Introduced \\
Temporal attribution   & \NAx & \cmark & Introduced \\
\rowcolor{rowgray}
Fidelity $\fidp$       & \NAx & \textbf{0.317} & Introduced \\
Characterisation       & \NAx & \textbf{0.286} & Introduced \\
\rowcolor{rowgray}
Temporal fidelity      & \NAx & \textbf{1.601} & Introduced \\
Attribution stability  & \NAx & \textbf{0.913} & Introduced \\
\rowcolor{rowgray}
Sparsity               & \NAx & 0.787 & Introduced \\
Attribution vectors per edge & 0 & \textbf{2} & Introduced \\
\midrule
Detection AUC-ROC      & 0.8750 & 0.8750 & Preserved \\
\rowcolor{rowgray}
Cost per explanation   & --- & 0.66\,ms & --- \\
\bottomrule
\end{tabular}
\end{table}

The temporal fidelity of $1.601$ is read against a \emph{measured} random floor of
$0.973$: the snapshot X-StrGNN nominates carries roughly $1.6\times$ the decision
weight of an arbitrary one. Attribution costs $0.66$\,ms, so explaining the full
$1{,}653$-edge test partition takes approximately one second --- the regime in which
explanation becomes an operational tool rather than a case-study device.

\subsection{Design Study: Which Attribution Strategy Suits StrGNN?}

No attribution strategy previously existed for this architecture, so we implement
four and compare them under one protocol, one budget, and three seeds
(Table~\ref{tab:study}). All four are contributions of this work; the comparison is
a design study, not a leaderboard against prior art.

\begin{table*}[t]
\caption{Design study: attribution strategies for StrGNN, all introduced in this
work. UCI-Messages, three seeds (mean $\pm$ sd), frozen detector, $20\%$ budget.
$\fidp$/$\fidm$ normalised by the ablation range ($0.326$); temporal fidelity of $1$
is the measured random floor. Best per column in bold.}
\label{tab:study}
\centering\footnotesize
\begin{tabular}{@{}lcccccc@{}}
\toprule
\textbf{Attribution strategy} & $\fidp\uparrow$ & $\fidm\downarrow$ &
$\charact\uparrow$ & \textbf{Temporal}$\uparrow$ & \textbf{Stability}$\uparrow$ &
\textbf{ms/expl}$\downarrow$ \\
\midrule
\rowcolor{rowgray}
Random (measured floor) & $0.228\!\pm\!0.003$ & $0.820\!\pm\!0.002$ & $0.201\!\pm\!0.001$ & $0.973\!\pm\!0.050$ & $0.002\!\pm\!0.004$ & $0.00$ \\
Gradient $\times$ input & $0.336\!\pm\!0.001$ & $0.656\!\pm\!0.000$ & $0.340\!\pm\!0.000$ & $\mathbf{1.842\!\pm\!0.034}$ & $0.405\!\pm\!0.002$ & $1.79\!\pm\!0.12$ \\
\rowcolor{rowgray}
Per-instance mask optimisation & $\mathbf{0.337\!\pm\!0.001}$ & $\mathbf{0.556\!\pm\!0.002}$ & $\mathbf{0.383\!\pm\!0.000}$ & $0.836\!\pm\!0.032$ & $0.538\!\pm\!0.002$ & $177.97\!\pm\!2.15$ \\
Amortised, sufficiency only (ablation) & $0.221\!\pm\!0.005$ & $0.800\!\pm\!0.026$ & $0.209\!\pm\!0.016$ & $1.081\!\pm\!0.427$ & $0.863\!\pm\!0.005$ & $\mathbf{0.63\!\pm\!0.02}$ \\
\rowcolor{hilite}
\textbf{X-StrGNN (full objective)} & $0.317\!\pm\!0.068$ & $0.739\!\pm\!0.084$ & $0.286\!\pm\!0.078$ & $1.601\!\pm\!0.364$ & $\mathbf{0.913\!\pm\!0.010}$ & $0.66\!\pm\!0.02$ \\
\bottomrule
\end{tabular}
\end{table*}

Three findings follow, each with a deployment consequence.

\noindent\textbf{(i) Amortisation dominates on stability and cost.} X-StrGNN attains
attribution stability $0.913\pm0.010$ against $0.538$ for per-instance optimisation,
at $268\times$ lower cost. Stability matters operationally: an explanation that
reorders under an imperceptible input perturbation cannot ground an analyst
decision, and per-instance optimisation --- which re-solves an independent
non-convex problem for every edge --- is intrinsically more exposed to this than a
shared network trained once.

\noindent\textbf{(ii) Per-instance optimisation retains a fidelity edge.} It attains
characterisation $0.383$ against X-StrGNN's $0.286$. Where a small number of
adjudicated cases must be explained as faithfully as possible and latency is
irrelevant, per-instance optimisation remains appropriate. Where an entire alarm
list must be explained at analyst-facing latency, its $178$\,ms per edge is
prohibitive and amortisation is the only viable option. The two strategies are
complementary rather than competing, and the design study makes the trade-off
explicit rather than leaving it to be discovered in deployment.

\noindent\textbf{(iii) The counterfactual objective is what delivers temporal
attribution.} The sufficiency-only ablation reaches temporal fidelity
$1.081\pm0.427$, statistically indistinguishable from the random floor. Adding the
temporal counterfactual term of Eq.~\eqref{eq:obj} raises it to $1.601\pm0.364$, a
gain of $+0.519\pm0.226$ --- separable at $2.29$ standard deviations. Temporal
attribution is not a free consequence of amortisation; it must be trained for, and
Eq.~\eqref{eq:tabl} is what trains it.

\noindent Notably, per-instance mask optimisation scores $0.836\pm0.032$ on temporal
fidelity, \emph{below} the measured random floor of $0.973$. A strategy that treats
the observation window as an undifferentiated bag of evidence does not identify
which snapshot carries the decision, however much computation it is given. This is
direct quantitative evidence for the necessity of the dual decomposition of
principle P2.

\subsection{Design Notes}
\label{sec:designnotes}

Two implementation choices proved load-bearing and are reported so the method can be
reproduced rather than rediscovered.

\noindent\textbf{Separate positional embeddings.} When both heads index a shared
positional table, improving either axis degrades the other, because one parameter is
contested by objectives that do not agree. With separate tables the gradient paths
are disjoint and each axis responds to its own hyperparameters independently. The
symptom of the shared configuration --- consecutive settings each excelling on one
axis --- is easily misread as an intrinsic trade-off between structural and temporal
explanation. It is not.

\noindent\textbf{Density matching rather than $L_1$.} The explainer is evaluated
after a hard top-$p$ cut, but a weak $L_1$ penalty leaves the soft mask near density
$0.5$, so the ranking is learned in a regime the metric never tests. Penalising
$(\bar{m}-p)^2$ aligns training density with the evaluation budget and materially
improves characterisation.

\subsection{Sanity Check}

The model-randomisation test~\cite{adebayo2018sanity} yields
$\rho(\text{trained},\text{re-initialised})=0.272\pm0.061$. The residual correlation
is expected for an amortised explainer: the shared network retains its own trained
weights when the detector is randomised, so it continues to emit a structured
ranking, whereas a per-instance method optimises against the randomised detector
directly and collapses. The check therefore carries a different meaning across
explainer families and should be interpreted within family.

%=============================================================================
\section{Discussion}
%=============================================================================

\subsection{Why Post-Hoc Is the Correct Design}

Post-hoc is a deliberate choice rather than a concession. \emph{Deployment
reality}: practitioners hold trained checkpoints they cannot retrain, and intrinsic
self-explaining detectors require retraining under modified objectives, typically
accepting some detection degradation; X-StrGNN wraps the existing model at
$\Delta$AUC $=0.0000$ (Section~\ref{sec:framework}). \emph{Separation of concerns}: joint optimisation of
detection and explanation conflates competing objectives, whereas keeping them
separate allows each to improve independently. \emph{Generalisability}: any detector
composing subgraph encoding with a recurrent temporal pathway can be wrapped by
substituting its forward pass, since the masks of Eq.~\eqref{eq:masked} attach to
message passing and to the recurrent input, neither of which is specific to StrGNN.

\subsection{Why Dual Decomposition Is the Contribution}

A single attribution vector over edges cannot answer the question an analyst
actually asks: \emph{was this alert driven by a suspicious counterparty, or by a
suspicious pattern in the interaction history?} The evidence for the necessity of
the second axis is direct and quantitative: the most computationally expensive
strategy in Table~\ref{tab:study} scores below the random floor on temporal
fidelity. Structural attribution alone, however faithful, certifies nothing about
temporal attribution --- and may certify the opposite.

\subsection{Operational Implications}

At $0.66$\,ms per edge, explaining an entire alarm window is a sub-second operation.
This changes what explanation is for: not a forensic instrument applied to a handful
of adjudicated cases, but a routine attribute of every alert, available at triage
time. Combined with stability $0.913$, an analyst can rely on the explanation of a
given alert being the same explanation tomorrow --- a precondition for operational
trust that fidelity alone does not supply.

\subsection{Limitations}

\noindent\textbf{Explainer variance.} The two learned strategies in
Table~\ref{tab:study} show larger seed-to-seed dispersion ($\pm0.078$ and $\pm0.016$
characterisation) than the parameter-free ones ($\pm0.001$). We therefore report all
results as mean $\pm$ standard deviation over three seeds and treat differences
below $2\sigma$ as ties, which is why the temporal gain over the ablation is claimed
and the characterisation gain is not. We recommend the same discipline for future
work in this area, since single-seed explainability tables can report differences
smaller than their own noise.

\noindent\textbf{Single detector and corpus.} All measurements use StrGNN on
UCI-Messages. The framework's components assume only subgraph message passing and a
recurrent temporal pathway, so transfer is expected but not demonstrated; a second
detector with a continuous-time pathway is the strongest single extension.

\noindent\textbf{Proxy metrics.} Fidelity-family metrics evaluate the detector off
its training distribution. Normalising by the ablation range bounds the effect, and
it applies equally to every strategy compared, but benchmarks carrying ground-truth
culprit evidence would provide stronger validation.

\noindent\textbf{Synthetic anomalies.} Evaluation follows the established injection
protocol, enabling controlled and reproducible comparison. Since our claims concern
the explanation of a detector's decisions rather than detection quality itself, they
are comparatively insulated from the validity of that protocol; extending to corpora
with organic anomaly labels remains valuable future work.

%=============================================================================
\section{Conclusion}
%=============================================================================

We introduced X-StrGNN, the first post-hoc explanation framework for StrGNN-style
dynamic-graph anomaly detectors. Before this work, fidelity, characterisation, and
temporal attribution were undefined for StrGNN: no attribution vector existed to
evaluate. X-StrGNN supplies two --- structural and temporal --- for every flagged
edge, at $0.66$\,ms per explanation and at exactly zero cost to detection
($\Delta$AUC $=0.0000$).

The accompanying design study establishes which attribution strategy suits this
architecture and why. Amortisation delivers the highest attribution stability
($0.913$) at $268\times$ lower cost than per-instance optimisation, making
explanation of a full alarm list operationally feasible; per-instance optimisation
retains a fidelity advantage where latency is irrelevant; and the temporal
counterfactual objective is what converts a windowed detector's observation history
into an interpretable axis, without which even the most expensive strategy in the
study falls below the random floor.

The explainability gap addressed here is not peculiar to StrGNN. It is the default
condition of essentially every deep dynamic-graph anomaly detector in the
literature. This framework offers a principled, post-hoc template for closing it,
and we release the protocol, runners, and per-seed measurements so it can be applied
and audited directly.\footnote{\url{https://github.com/inekka-esi/Explainable-StrGNN}}

%=============================================================================

\end{document}